\documentclass[draftclsnofoot, peerreviewca]{IEEEtran}
\usepackage{
	amsmath,%
	amsfonts,%
}
\usepackage{url}
\usepackage{graphicx}
\usepackage[hidelinks]{hyperref}	
\usepackage{siunitx}  
\usepackage{tikz}
\usetikzlibrary
{%
	patterns,
	positioning,
	decorations.pathreplacing,
}%

\input{99_custommacros.sty}
\newcommand{\timesteptxt}{time~step~\(\timeindex\)\xspace}
\newcommand{\timestepstxt}{time~steps~\(\timeindex\)\xspace}
\newcommand{\jobtxt}{job~\(\jobindex\)\xspace}
\newcommand{\jobstxt}{jobs~\(\jobindex\)\xspace}

\newcommand{\internalprobability}					{p}
\newcommand{\probjob}{\internalprobability_{\text{job}}}
\newcommand{\probprio}{\internalprobability_{\text{prio}}}

\newcommand{\internalpercentage}					{a}
\newcommand{\percentageuser}[1][\userindex]{\internalpercentage_{#1}}

\newcommand{\totalusers}							{N}
\newcommand{\userindex}								{n}

\newcommand{\jobindex}								{j}
\newcommand{\internalsetjob}						{\mathbb{J}}
\newcommand{\setjobsuser}[1][\userindex]{\internalsetjob_{{#1},  \timeindex}}

\newcommand{\totalresourceblocks}					{U}
\newcommand{\internalresourceblock}					{u}
\newcommand{\resourceblocksrequesteduser}[1][\jobindex]{\internalresourceblock_{#1, \text{req}, \timeindex}}

\newcommand{\resourceblocksinitialuser}{\internalresourceblock_{\userindex,\text{init},\timeindex}}
\newcommand{\resourceblocksinitialmax}{\internalresourceblock_{\text{max}}}
\newcommand{\resourcesscheduleduser}{\internalresourceblock_{\userindex,\text{sx},\timeindex}}

\newcommand{\internaldelay}							{d}
\newcommand{\delayjob}{\internaldelay_{\jobindex,\timeindex}}
\newcommand{\delaymax}{\internaldelay_{\text{max}}}

\newcommand{\channelfadingamplitude}{|\internalchannelcoefficient_{\userindex,\timeindex}|}
\newcommand{\channelfadingpower}{\channelfadingamplitude^{\num{2}}}
\newcommand{\rayleighscale}{\variance_{\text{R}}}

\newcommand{\reward}{\internalreward_{\timeindex}}
\newcommand{\rewardestim}{\hat{\internalreward}_{\timeindex}}
\newcommand{\rewardtimeoutnormal}{\internalreward_{\internaldelay,\timeindex,\text{normal}}}
\newcommand{\rewardtimeoutprio}{\internalreward_{\internaldelay,\timeindex,\text{prio}}}
\newcommand{\rewardcapacity}{\internalreward_{c,\timeindex}}
\newcommand{\weighttimeoutnormal}{\internalweight_{\internaldelay,\text{normal}}}
\newcommand{\weighttimeoutprio}{\internalweight_{\internaldelay,\text{prio}}}
\newcommand{\weightcapacity}{\internalweight_{c}}

\newcommand{\actionvec}{\mathbf{\internalaction}_{\timeindex}}
\newcommand{\actionvecnoise}{\mathbf{\tilde{\internalaction}}_{\timeindex}}
\newcommand{\statevec}{\mathbf{\internalstate}_{\timeindex}}
\newcommand{\statescauser}[1]{\internalstate_{\userindex,{#1},\timeindex}}

\newcommand{\criticnetwork}{\internalcritic ( \statevec,  \actionvec,  \parameterscritictime ) }
\newcommand{\actornetwork}{\boldsymbol{\internalactor} ( \statevec,  \parametersactortime )}

\newcommand{\parameterindex}						{i}
\newcommand{\totalparameters}						{I}
\newcommand{\internalfisherinformation}				{F}
\newcommand{\parametertime}[1][\parameterindex]{\internalparameter_{#1,  \timeindex}}
\newcommand{\parametercritical}[1][\parameterindex]{\internalparameter_{\parameterindex}^{\text{Anchor}}}
\newcommand{\parameters}{\boldsymbol{\internalparameter}}
\newcommand{\parametersactor}{\parameters_{\internalactor}}
\newcommand{\parametersactortime}{\parameters_{\internalactor,  \timeindex}}
\newcommand{\parameterscritictime}{\parameters_{\internalcritic,  \timeindex}}

\newcommand{\fisherinformationcritical}{\internalfisherinformation_{\parameterindex}^{\text{Anchor}}}
\newcommand{\fisherinformationtime}[1][\timeindex]{\internalfisherinformation_{\parameterindex,{#1}}}
\newcommand{\explorationmomentum}{\internalexplorationparam_{\text{expl}}}

\newcommand{\internalloss}							{L}
\newcommand{\losscritic}{\internalloss_{\internalcritic,\timeindex}}
\newcommand{\lossactor}{\internalloss_{\internalactor,\timeindex}}
\newcommand{\lossanchor}{\internalloss_{\parametercritical,\timeindex}}
\newcommand{\weightanchor}{\internalweight_{\parametercritical}}

\begin{document}

\title{Robust Deep Reinforcement Learning Scheduling via Weight Anchoring}

\author{%
	Steffen Gracla%
	,
	Edgar Beck%
	,
	Carsten Bockelmann%
	\ and
	Armin Dekorsy%
	\thanks{This work was partly funded by the German Ministry of Education and Research (BMBF) under grant 16KIS1028 (MOMENTUM).}%
	\thanks{All authors are with the Department of Communications Engineering, University of Bremen, Germany. Email: \{gracla, beck, bockelmann, dekorsy\}@ant.uni-bremen.de}%
	\thanks{This work was accepted for publication in IEEE Communications Letters.}
}

\markboth{%
	IEEE Communications Letters,~Vol.~x, No.~y, Month~0000}%
	{Shell \MakeLowercase{\textit{et al.}}: A Sample Article Using IEEEtran.cls for IEEE Journals%
}

\IEEEpubid{0000--0000/00\$00.00~\copyright~0000 IEEE}

\maketitle

\begin{abstract}
	Questions remain on the robustness of data-driven learning methods when crossing the gap from simulation to reality.
	We utilize weight anchoring, a method known from continual learning, to cultivate and fixate desired behavior in Neural Networks.
	Weight anchoring may be used to find a solution to a learning problem that is nearby the solution of another learning problem.
	Thereby, learning can be carried out in optimal environments without neglecting or unlearning desired behavior.
	We demonstrate this approach on the example of learning mixed QoS-efficient discrete resource scheduling with infrequent priority messages.
	Results show that this method provides performance comparable to the state of the art of augmenting a simulation environment, alongside significantly increased robustness and steerability.
\end{abstract}




\section{Introduction}
\label{sec:introduction}

{%
	In the field of communication systems,
}%
deep Reinforcement Learning~(\reinforcementlearning) has stirred interest with its ability to learn approximately optimal strategies with limited model assumptions.
This is an auspicious promise for challenges that are complex to model or to solve in real time, and early research has shown that deep \reinforcementlearning strategies are indeed applicable to problems such as intelligent resource scheduling~\cite{zhang_survey_2019, gracla_ddpg_2022}.
{%
	Ultra-reliability, low latency, and heterogeneous QoS-constraints are cornerstones of future communication systems, particularly in fields such as medical communications, and flexible and fast learned schedulers may aid in achieving required performance goals.
}
However, to answer whether these learned strategies can truly uplift modern communication systems, key questions regarding the \emph{reliability} of these learned strategies in highly demanding scenarios must be addressed.
Robustness and sample efficiency are among the central issues of Deep Learning~(\deeplearning) that may cast doubt on its viability in the near future.
For example, the commonly used deep deterministic policy gradient~(\ddpg) algorithm is known to have issues with sparse events~\cite{matheron_problem_2019, fujimoto_addressing_2018}.

These issues are commonly addressed by making a virtue of the sample-hunger of \deeplearning.
Training data are often generated synthetically, so one may \emph{augment} the simulation environment that generates the data by, \eg increasing the occurrence of rare events like emergency signals~\cite{kasgari_experienced_2020} or splicing relevant examples into the training data~\cite{chae_autonomous_2017}.
{%
	While this approach is usable, highly parametrized training environments may require time-intensive tweaking to produce desired behavior on the learned algorithm.
}
Worse, this approach widens the gap between simulation environment and reality, which is known to impede Machine Learning~\cite{pinto_robust_2017}.
Finally, this approach does not inherently prevent a phenomenon known as ``catastrophic forgetting''~\cite{kirkpatrick2017overcoming}, where already learned behavior may be overwritten if examples are not encountered frequently.

{%
	In this paper, we regard the problem of discrete resource scheduling with rare but important \urllc priority messages.
	We make use of generic deep \reinforcementlearning algorithms, but propose to cast the handling of priority signals and the handling of normal scheduling as two distinct tasks.
	By doing this, we enable the use of a method from the continual learning domain, \emph{weight anchoring}, as discussed in~\cite{kirkpatrick2017overcoming}.
	Motivated by information theory, this method uses the Fisher information to focus the optimization landscape on solutions in the vicinity of the anchored solution via an elastic penalty.
	Hence, we design a two-stage learning process.
	First, our approach learns to handle priority messages exclusively.
	Next, we apply weight anchoring to this solution, and learn to optimize for overall system performance.
}

Such a split-task approach promises certain procedural advantages: Both priority handling and normal scheduling can be learned without distraction on environments that offer a low reality gap for optimal, sample-efficient learning.
A single scaling factor controls the anchored tasks pull on the main optimization objective.
Additionally, approximations of the Fisher information are readily available at no additional computation cost, as modern gradient descent optimizers already make use of it.
{%
	In this work, we compare the commonly used ``augmented simulation'' approach with our two-stage process.
	We show comparable results in both overall system performance and handling of priority messages.
	We further highlight that our process is affected significantly less by catastrophic forgetting.
}





\section{Setup \& Notations}
\label{sec:setup}
In this section, we introduce the general setup of the scheduling task and its metrics.
We give detail on the standard actor-critic deep \reinforcementlearning algorithm that is used to find efficient allocation solutions,
and the elastic weight anchoring which encodes priority task handling.
\IEEEpubidadjcol

\subsection{The Allocation Task}
\label{sec:allocation_task}

\begin{figure}[!t]
	\centering
	\includegraphics[scale=0.99]{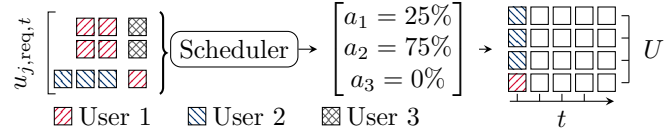}
	\caption{%
		The Resource allocation scenario tasks a scheduler with assigning a fraction of given discrete resources \( \totalresourceblocks \) to each user \( \userindex \) at each time step~\(\timeindex\).
		It schedules based on the current state of a request pool, with request jobs of varying sizes and time-outs assigned to one particular user.
	}
	\label{fig:scenario}
\end{figure}

Many current medium access schemes, such as Orthogonal Frequency-Division Multiplexing~(\ofdm), assume division of the total available resource bandwidth into discrete blocks.
Hence, we introduce a scenario as depicted in Fig.~\ref{fig:scenario}, where in each discrete time~step~\( {\timeindex \in \numbersnatural} \) a scheduler is tasked with assigning the limited number~\( {\totalresourceblocks \in \numbersnatural} \) of discrete resource blocks to jobs~\( {\jobindex \in \numbersnatural} \).
Jobs~\( \jobindex \) come in two types, \emph{normal} and \emph{priority}, to be explained subsequently, and have two attributes: 1)~A request size~\( {\resourceblocksrequesteduser \in \numbersnatural} \) in resource blocks; 2)~A delay counter~\( {\delayjob \in \numbersnatural} \) in \timestepstxt since the \jobtxt has been generated.
Job generation occurs at the beginning of each \timesteptxt at a probability~\( \probjob \) for each of the~\( \totalusers \in \numbersnatural \) connected users.
When generated, a \jobtxt is assigned an initial size~\( {\resourceblocksrequesteduser \gets \resourceblocksinitialuser \sim \mathbb{U}\left[\num{1},\,\resourceblocksinitialmax \right]} \) drawn from a discrete uniform distribution, and the delay is initialized as \( \delayjob \gets \num{0} \).
In a \timesteptxt, the scheduler outputs allocations~\( \actionvec \) that distribute a ratio of the total resources~\( \totalresourceblocks \) to each of the users~\( \userindex \),
\begin{align}
	\actionvec
	=
		\left[
			\percentageuser[1], \quad
			\percentageuser[2], \quad
			\dots, \quad
			\percentageuser[\totalusers]
		\right]
	,
	\quad
	\text{with}\ 
	\sum_{\userindex = \num{1}}^{\totalusers} \percentageuser = \num{1}
	.
\end{align}
According to this allocation vector, the discrete blocks are then distributed to jobs currently in queue, starting from the oldest \jobtxt of each user~\( \userindex \).
The job's requested blocks~\( \resourceblocksrequesteduser \) are decreased accordingly.
Once all resources have been distributed following an allocation~\( \actionvec \), the delay~\( \delayjob \) of all \jobstxt with requests remaining, \ie~\( {\resourceblocksrequesteduser > \num{0}} \), is incremented.
Jobs~\( \jobindex \) with a delay~\( {\delayjob > \delaymax} \) greater than a maximum allowed delay~\( \delaymax \) at the end of a \timesteptxt are removed from the queue, and a count~\( \rewardtimeoutnormal \) of timed-out jobs in that \timesteptxt is incremented.
Further, at the beginning of each \timesteptxt there is a probability~\( \probprio \) that one existing \jobtxt is designated \emph{priority} status.
If a priority job has not been fully scheduled within one \timesteptxt, it is removed from the job queue regardless of its delay~\( \delayjob \), incrementing a count~\( \rewardtimeoutprio \) of timed-out priority jobs in \timesteptxt.

In addition to minimizing time-outs~\( \rewardtimeoutnormal \) and, in particular, \( \rewardtimeoutprio \), the scheduler will also be tasked with maximizing the sum-capacity of communication.
The connection between base-station scheduler and users~\( \userindex \) is described by a Rayleigh-fading channel with a fading amplitude~\( {\channelfadingamplitude \sim \text{Rayleigh}(\rayleighscale)}\) drawn from a Rayleigh distribution with scale~\( \rayleighscale \) in each \timesteptxt.
Unlike in usual \ofdm systems, we assume the power fading to be independent of the specific resource selected, to reduce computation complexity within the scope of this work.
With \( \resourcesscheduleduser \) being the resources scheduled to a user~\( \userindex \) in a \timesteptxt, the sum capacity under Gaussian code books calculates as
\begin{align}
	\label{eq:capacity}
	\rewardcapacity
	=
		\sum_{\userindex = \num{1}}^{\totalusers}
		\resourcesscheduleduser
		\cdot
		\log
		\left(
			\num{1}
			+
			\channelfadingpower
			\frac{\signalpower}{\noisepower}
		\right),
\end{align}
where we assume the ratio of expected signal power~\( \signalpower \) and expected noise power~\( \noisepower \) as constant for all users.
We aim for the scheduler to balance these three goals~\( \rewardtimeoutnormal \), \( \rewardtimeoutprio \), and \( \rewardcapacity \) with a weighting of our choosing.
Therefore, we collect all metrics in a reward sum
\begin{align}
	\label{eq:reward}
	\reward
	=
		\weightcapacity\,\rewardcapacity
		- 
		\weighttimeoutnormal \,\rewardtimeoutnormal
		-
		\weighttimeoutprio \,\rewardtimeoutprio
\end{align}
to be maximized, with tunable weights~\( \weightcapacity \), \( \weighttimeoutnormal \), \( \weighttimeoutprio \) that control the relative importance of each term.

\subsection{Deep Reinforcement Learning-Allocation}
\label{sec:drlallocation}

We implement a \ddpg-based scheduler, as used in our previous work~\cite{gracla_ddpg_2022}, to learn to output allocations~\( \actionvec \) that approximately optimize the sum reward~\( \reward \) in \refeq{eq:reward}.
The scheduler uses a pre-processor, two neural networks (\emph{actor} and \emph{critic}), a memory module, an exploration module, and a learning module.
{%
	In the following we will briefly describe these components and the scheduler's process flow in training and inference.
	This scheduler has no explicit way of dealing with potential sparsity of priority job events.
}

We begin the design of our scheduler with a pre-processor that summarizes the information relevant to making allocation decisions into a vector of fixed size, palatable for the neural networks that are used subsequently.
This state vector~\( \statevec \) holds four features per user~\( \userindex \):
\begin{enumerate}
	\item Resources requested in the set~\( \setjobsuser \) of jobs assigned to user~\( \userindex \) at \timesteptxt, normalized by total resources~\( \totalresourceblocks \):
	\(
		\statescauser{\num{1}}
		=
			\sum_{\jobindex \in \setjobsuser}
			\resourceblocksrequesteduser \,
			/ \,
			\totalresourceblocks
	\)
	\item Priority resources requested, normalized by total resources~\( \totalresourceblocks \):
	\(
		\statescauser{\num{2}}
		=
			\sum_{\jobindex \in \{\setjobsuser,\, \jobindex \text{ is prio}\}}
			\resourceblocksrequesteduser \,
			/ \,
			\totalresourceblocks
	\)
	\item Instantaneous channel power fading:
	\(
		\statescauser{\num{3}}
		=
			\channelfadingpower
	\).
	{%
		We assume the power fading to be perfectly estimated; in a real system, an error or time delay may interfere.
	}
	\item Maximum delay, normalized by maximum allowed delay:
	\(
		\statescauser{\num{4}}
		=
			\max_{\jobindex \in \setjobsuser}
				\delayjob \,
				/ \,
				\delaymax
	\)
\end{enumerate}
{%
	The state is therefore directly influenced by the allocation action \( \actionvec \) taken by the scheduler in the previous \timesteptxt.
}
Observed states~\( \statevec \) are saved in the memory module to be used by the learning module later, and fed into the \mbox{\emph{actor}-network~\( \actornetwork \)}.
Based on the state and its current parameters~\( \parametersactortime \), the actor outputs an allocation~\( \actionvec \).
In order to experience a wide variety of state-action combinations, an exploration module then introduces noise to the action~\( \actionvec \) as follows.
Using a momentum parameter~\( \explorationmomentum \) that is decayed to zero over the course of training, the actor allocation~\( \actionvec \) is mixed with a normalized vector~\( \actionvecnoise \) of same length with entries drawn from a random uniform distribution~\( {U(\num{0}, \num{1})} \), as \({\actionvec \gets \explorationmomentum \actionvecnoise + (\num{1} - \explorationmomentum) \actionvec}\).
The noisy action is then re-normalized and propagated to the greater communication system, where success metrics are calculated according to \refsec{sec:allocation_task}.
Both noisy action and the resulting reward~\( \reward \) are saved into the memory buffer alongside their state~\( \statevec \), and a new, updated state is forwarded to the pre-processor, restarting the loop.

In order to learn how to output good allocations from the experiences made, the learning algorithm must answer the question: What is a \emph{good} mapping of state vector~\( \statevec \) to allocation~\( \actionvec \)?
Standard Deep Deterministic Policy Gradient~(\ddpg)~\cite{lillicrap2015continuous} algorithms decompose this question into two separate, fully connected neural networks with learnable parameters~\(
	{\parameterscritictime, \parametersactortime}
\), respectively.
As described previously, the actor-network handles the direct mapping of a state to an allocation.
Meanwhile, the \emph{critic}-network~\( { \criticnetwork = \rewardestim } \) identifies what makes an allocation ``good'' by estimating the expected success~\( \reward \), \ie~\refeq{eq:reward}, of performing an allocation~\( \actionvec \) in state~\( \statevec \).
It therefore approximates the unknown dynamics of the system that lead from allocation~\( \actionvec \) to metric~\( \reward \), \ie in this case: 1)~the timeout mechanic; 2)~the sum capacity formula in (\ref{eq:capacity}); and 3)~the relative weightings~\( \internalweight \) in (\ref{eq:reward}).
While in the learning phase, once per \timesteptxt, both networks learn from the data set of experiences collected in the memory module as follows.
A mini-batch of experiences~\( (\statevec, \actionvec, \reward) \) is sampled from the memory module and used to evaluate a loss function for critic and actor each.
The critic loss provides the squared error between critic estimate~\( \criticnetwork \) and experienced reward~\( \reward \):
\begin{align}
	\losscritic
	= 
		\left(
			\criticnetwork
			-
			\reward
		\right)^{\num{2}}
	.
\end{align}
Thus, a lower loss~\( \losscritic \) corresponds to a better reward estimate.
Meanwhile, the \emph{actor}-loss evaluates the critics reaction to the actors allocation:
\begin{align}
	\label{eq:lossactorvanilla}
	\lossactor
	=
		-
		\internalcritic
		\left(
			\statevec, \,
			\actornetwork, \,
			\parameterscritictime
		\right)
	.
\end{align}
A lower loss~\( \lossactor \) indicates greater estimated rewards.
Each networks' parameters are then tuned using variants of stochastic gradient descent~(\stochasticgradientdescent), leveraging the gradients~\( {\nabla_{\parameters} \internalloss} \) to minimize the respective loss functions.
The actor’s allocation strategy is therefore adjusted in a way that maximizes rewards~\( \rewardestim \), as estimated by the critic.
{%
	These parameter updates can be performed decoupled from inference and sample gathering.
	In run-time inference, the process flow simplifies to only the pre-processor and the actor network.
}

\subsection{Weight Anchoring}
\label{sec:weightanchoring}

In our split-task approach, we are looking to first train the scheduler to deal with priority messages, then preserve this behavior as we learn the greater objective of QoS-efficient scheduling.
A neural network's behavior is determined by its parameters~\( \parameters \).
Thus, after the scheduler has learned to deal with priority messages to satisfaction in learning stage one, we follow~\cite{kirkpatrick2017overcoming} and record two factors for each parameter~\( \parameterindex \): 1) The final values~\( \parametercritical \) describe a parametrized network that deals well with priority messages; 2) the Fisher information~\( \fisherinformationcritical \) measures how sensitive the network behavior reacts to local changes on parameter~\( \parameterindex \).
For a more in depth analysis of Fisher information in the context of neural networks we refer to~\cite{achille2018critical}.
To save on computation cost, we extract an approximation of the Fisher information used by the \stochasticgradientdescent-optimizer.
In our case, the Adam optimizer estimates the Fisher information as a moving average of the squared gradient~\cite{kingma2014adam},
\begin{align}
	\fisherinformationtime
	\gets
	\frac{
		\beta_{\num{2}}
		\cdot
		\fisherinformationtime[\timeindex - \num{1}]
		+
		\left(
		\num{1} - \beta_{\num{2}}
		\right)
		\left(
		\nabla_{\parametersactor}
		\lossactor
		\right)^{\num{2}}
	}
	{
		\left(
		\num{1} - \beta_{\num{2}}
		\right)
	}
	,
\end{align}
with~\( \beta_{\num{2}} \) controlling the momentum of the rolling average in the optimizer.

Now, in learning stage two, we would like to learn QoS-efficient scheduling on the same network or any network of the same dimensions without straying from the critical message handling learned in stage one.
To do this, we can use \( \parametercritical \) and \( \fisherinformationcritical \) to construct a term \( \lossanchor \) that measures how far this network's current behavior, represented by its current parameter values~\( \parametertime \), has moved from the recorded behavior:
\begin{align}
	\label{eq:lossanchor}
	\lossanchor = 
	\weightanchor
	\sum_{\parameterindex=\num{1}}^{\totalparameters}
	\fisherinformationcritical
	\left(
	\parametertime
	-
	\parametercritical
	\right)^{\num{2}}
	.
\end{align}
This term grows proportional to the distance to the recorded parameter values~\( \parametercritical \), weighted by their sensitivity~\( \fisherinformationcritical \), and is scaled by an anchoring weight~\( \weightanchor \).

By adding this term \( \lossanchor \) to the actor loss \refeq{eq:lossactorvanilla}, we can incentivize the actor to learn a parametrization~\( \parametersactor \) that not only maximizes the approximated rewards, but also minimizes the distance to the recorded parameters~\( \parametercritical \).
The anchoring weight~\( \weightanchor \) controls how strongly the term \refeq{eq:lossanchor} pulls the learning process to solutions in the vicinity.



\section{Experiments}
\label{sec:experiments}
In an environment with priority messages whose frequency is controlled by a parameter \( \probprio \), we present and compare the performance of different \deeplearning schedulers, {%
	both} with and without anchoring. While the {%
	true} baseline frequency of these priority messages is assumed to be rare ({\( \probprio = \SI{0.01}{\percent} \)}), some schedulers will learn in an \emph{augmented} environment with artificially increased priority message frequency. For our anchored schedulers, we also show the impact for different choices of anchoring weight~\(\weightanchor\).

\subsection{Implementation Details}
\label{sec:implementationdetails}

We implement the simulation using Python and the TensorFlow library using the Adam~\cite{kingma2014adam} optimizer. The full implementation code is provided in~\cite{gracla2021code}, and the system configuration during network training is listed in \reftab{tab:trainingconfiguration}.
We assume the reward weights \( \internalweight \) and channel design parameters to be tuned by an expert.
All schedulers' performance is evaluated on the baseline environment with rare ({\( \probprio = \SI{0.01}{\percent} \)}) priority messages. In total, we train eight different \deeplearning schedulers:

1) The baseline scheduler~(BS) is trained exclusively on the same environment that all schedulers will be evaluated on, \ie priority messages appear infrequently~(\( {\probprio = \SI{0.01}{\percent}} \)).

2) Another scheduler~(AU20) is trained on an augmented simulation environment where priority messages are encountered much more frequently~(\( {\probprio = \SI{20}{\percent}} \)).

3) For the first stage of our anchoring approach we train a scheduler~(AU100) to deal well with priority messages~(\( {\probprio = \SI{100}{\percent}} \)). We later evaluate a snapshot of the scheduler at this point to highlight the performance gap.

4-6) For the second stage of our anchoring scheduler, we initialize three networks with the weights learned by scheduler no.~3, AU100. We then train these schedulers (AN1, AN2, AN3) on the baseline simulation~(\( {\probprio = \SI{0.01}{\percent}} \)), while anchored to AU100's weights as described in \refsec{sec:weightanchoring}. AN1, AN2 and AN3 differ in choice of the anchoring parameter~\( \weightanchor \), listed in \reftab{tab:trainingconfiguration}, to highlight its influence.

As our anchored schedulers are trained in two phases, with \( {\num{30} \times \num{10000}} \) training steps each, we double the training episodes for the non-anchored benchmark scheduler AU20 to normalize training time. Further,

7-8) To highlight the effect of forgetting, we initialize two networks~(AU20+, AN1+) with the parameters learned by no.~2, AU20, and no.~4, AN1, respectively. Both receive another stage of training on an environment with no (\( {\probprio = \SI{0}{\percent}} \)) priority messages. Scheduler AN1+ remains anchored to the priority scheduling solution AU100.

\begin{table}[!t]
	\renewcommand{\arraystretch}{1.0}
	\caption{Training Configuration}
	\label{tab:trainingconfiguration}
	\centering
	\begin{tabular}{ll|ll}
		\hline
		Steps~\( \timeindex \) per episode & \( \num{10000} \) 					&  Episodes & \( \num{30} \)
		\\
		SNR & \( \SI{10}{\decibel} \)											&  Rayleigh Scale~\( \rayleighscale \) & \( \num{0.3} \)
		\\
		Resource Blocks~\( \totalresourceblocks \) & \( \num{10} \) 			&  Users~\( \totalusers \) & \( \num{5} \)
		\\
		Max RB~\( \resourceblocksinitialmax \) & \( \num{7} \) 					&  Max Delay~\( \delaymax \) & \( \num{5} \) 
		\\
		Job Probability~\( \probjob \) & \( \SI{50}{\percent} \)  				&  Batch Size & \( \num{256} \)
		\\
		F. Mom. \( \beta_{\num{2}} \)& default 									&  NN Layers \( \times \) Nodes & \(\num{3} \times \num{128}\)
		\\
		Weight~\( \weightcapacity \) & \( +\num{1} \) 							&  Weight~\( \weightanchor \) & \( 1e\left\{ 5, 6, 7 \right\} \)
		\\
		Weight~\( \weighttimeoutnormal \) & \( -\num{1} \) 						&  Weight~\( \weighttimeoutprio \) & \( \num{-5} \)
		\\
		Init.~\( \explorationmomentum \)& \( \num{1} \) 						&  Episodes \( {\explorationmomentum \rightarrow \num{0}} \) & \( \SI{50}{\percent} \)
		\\
		\hline
	\end{tabular}
\end{table}



\subsection{Results}
\label{sec:results}
\begin{figure}[!t]
	\centering
	\includegraphics[scale=0.99]{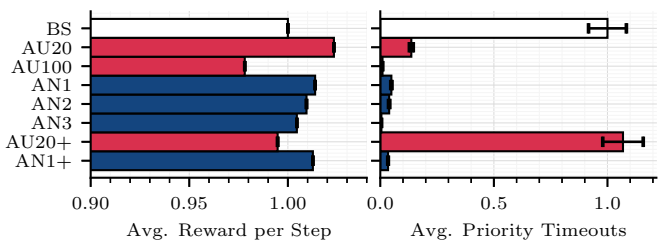}  
	\caption{Results for the eight schedulers introduced in \refsec{sec:implementationdetails}. We group schedulers by color, where the baseline scheduler BS is colored white, schedulers from augmented simulations are red, and anchored schedulers are blue. Results are normalized to the baseline scheduler, BS. For rewards, higher results are better, while for timeouts, lower results are better. Black bars represent the variance in results over all simulation runs.}
	\label{fig:results1}
\end{figure}

After training, all eight schedulers from \refsec{sec:implementationdetails} are frozen and evaluated on the setting with rare priority events~(\( {\probprio = \SI{0.01}{\percent}} \)).
Testing the schedulers is carried out over \( \num{5} \) episodes at \( \num{200000} \) steps per episode for an expected \(\num{10}\) priority events per episode.
The process of training and testing is repeated three times for all schedulers, their average results and variance in overall performance and priority time-outs displayed in \reffig{fig:results1}. All results are normalized to the performance of scheduler~BS, trained exclusively on~\( {\probprio = \SI{0.01}{\percent}} \), which serves as a baseline.
Scheduler AU100, which was trained exclusively on priority events, highlights two important points: 1)~It is possible to drastically reduce priority timeouts compared to the baseline~BS; 2)~There is a QoS-sum performance gap compared to baseline~BS.
We see that both the augmented scheduler~AU20 as well as the anchoring approaches~AN are able to fill or even exceed the reward performance gap, while still providing significantly better handling of priority messages.
The degree to which these approaches trade-off reward performance and time-out performance is visibly influenced by the anchoring parameter~\( \weightanchor \) for the anchoring approaches, and by the choice of priority message frequency~\( \probprio \) within the augmented simulation for the augmented scheduler.
{%
	We expect these trade-offs to scale nonlinearly.}

As mentioned in \refsec{sec:introduction}, the anchoring approaches~AN additionally offer certain desirable benefits over simply augmenting the simulation, as done in AU20, leading to better robustness, sample-efficiency, and a lesser model-gap. We highlight one of these benefits, the increased robustness against forgetting, in the results of schedulers AU20+, AN1+. They continue the training of AU20, AN1, respectively, without encountering any priority events. We find that our anchoring approach is able to retain performance and priority handling, while the comparison method AU20 ``forgets'' priority handling entirely.






\section{Conclusions}
\label{sec:conclusions}
We present a way to transfer weight anchoring, a method from multi-task learning motivated by information theory, to robust deep \reinforcementlearning in the presence of significant rare events.
The handling of rare priority events is cast as a separate task.
Its learnings are subsequently used as a basis to find a overall QoS optimal resource scheduling solution in the neighborhood of handling priority tasks.
This method brings procedural advantages and is of particular interest in applications that suffer from high complexity, but require a highly deliberate, sample-efficient handling of rare critical events, which is traditionally a weakness of \deeplearning.
We demonstrate this on a deep \reinforcementlearning discrete resource scheduling task, though the method itself is not limited to this application.

\bibliographystyle{ref/IEEEtran}
{%
	%
	\bibliography{ref/IEEEabrv,ref/references}
}%

\end{document}